\documentstyle[aaai209]{article}

\begin{document}

\title{SATEN: An Object-Oriented Web-Based Revision and Extraction Engine}
\author{Mary-Anne Williams and Aidan Sims\\
Business Technology Research Laboratory\\ 
University of Newcastle
\\ NSW 2308, Australia 
\\ maryanne@cafe.newcastle.edu.au \ \ \ \ \ \ \ \ \ aidan@math.newcastle.edu.au 
 }
\maketitle

\section{General Information}

SATEN\footnote{SATEN a Sagacious Agent for Theory Extraction and revisioN is available at  http://cafe.newcastle.edu.au/saten} is an object-oriented web-based extraction and belief revision engine. It runs on any computer 
via a Java 1.1 enabled browser such as Netscape 4. SATEN performs belief revision based on the AGM approach \cite{agm}.  The extraction and belief revision reasoning engines operate on a user specified ranking of information. One of the features of SATEN is that it can be used to integrate mutually inconsistent commensuate rankings into a consistent ranking.

\section{General Description of the System}

SATEN has two reasoning engines:  a revision engine and an extraction engine. Both engines operate on a user-specified ranking of information.
 
The process of {\sl extraction} involves the recovery of a consistent ranking from an inconsistent ranking. Several strategies are available for this process. Extraction can be seen as a generalisation of Brewka's work on subtheories \cite{Brewka}. It is also related to Poole's work \cite{Poole,Dubois}. 

{\sl Revision} models the incorporation of new information (a belief) into a knowledge base.  The main difficulty in modeling this process is that the incoming information may be inconsistent with the agent's knowledge base. Consequently, the agent must determine the information to be relinquished before the new information can be accepted, if the agent is to maintain a consistent knowledge base. There are several constructive definitions of belief revision based on various preference relations over the agents knowledge. These preference relations attempt to capture the relative importance of information with respect to change. When an agent is forced to give up beliefs in order to accept some new information then, if presented with a choice, a response would be to choose to retain more important information, like system laws and integrity constraints, rather than contingent facts and default facts. 

\vskip 8pt
\noindent 
SATEN is capable of: 
\begin{itemize}
 \item Theory Extraction
 \item Iterated Belief Revision 
  \item Nonmonotonic Reasoning 
 \item Possibilistic Reasoning 
 \item Hypothetical Reasoning
\end{itemize}

Using combinations of these capabilities SATEN can perform information integration/fusion on mutually inconsistent commensurate knowledge bases and can calculate
Spohnian reasons \cite{SpohnB}. 

SATEN currently implements the  reasoning strategies below based on  a user specified ranking of beliefs (logical formulae). A {\em ranking} can be specified as mapping from beliefs to [1,2,3,...] in which case the most important information is given the lowest rank, i.e. rank is inversely proportional to importance. Alternatively, beliefs can be ranked using {\em degrees of belief} between 0 and 1 - the higher the degree the more important the information, in particular tautologies are assigned 1, whilst inconsistencies and nonbeliefs are given 0. SATEN can derive an ordinal ranking from a set of specified degrees automatically, by toggling one of the  menu settings.

\begin{itemize}
\item Standard Adjustment \cite{JELIA,IJCAI95,KR94} is based on the standard epistemic entrenchment construction \cite{gm} for Belief Revision. 
\item Maxi-Adjustment is based on maximal inertia \cite{IJCAI97}. Maxi-adjustment proceeds from the top of the ranking and moves down it rank by rank. At each rank it deletes all the beliefs that are inconsistent with other beliefs at that rank and above.
\item Hybrid Adjustment is a combination of standard adjustment and maxi-adjustment \cite{IJCAI97}. Adjustment is computationally ``easier" than maxi-adjustment, however the main shortcoming of adjustment is that it maintains beliefs only if there is an \emph{explicit reason to keep them}, whilst maxi-adjustment removes beliefs only if there is an \emph{explicit reason to remove them}. Hybrid adjustment performs an adjustment which computes  the core beliefs to be retained, and then it performs a maxi-adjustment which is used to regather as many  beliefs  as possible that were discarded during the adjustment phase.
 \item Global Adjustment is similar to a maxi-adjustment except that it takes all beliefs in the ranking into consideration when computing the minimally inconsistent  subsets, instead of proceeding rank by rank. It removes the least ranked beliefs "causing" the inconsistency.
\item Linear Adjustment is similar to maxi-adjustment except that it removes all the beliefs at ranks which contain incosisitencies with beliefs above them \cite{Nebel}. IF there is only one belief at each rank, then linear and maxi-adjustment are identical.
\item Quick Adjustment is identical to maxi-adjustment except that rather
than removing all beliefs it randomly chooses a minimal number of beliefs from each set of
inconsistent beliefs in this way the inconsistency is removed.
 \end{itemize}

 A number of examples are available via a dropdown menu which highlight the different behaviour of the strategies.

\section{Methodology}

An inherent difficulty that arises when trying to model world states using the predicate calculus is that of consistency. Under the rules of formal logic, anything is derivable from a contradiction, hence it is imperative that consistency of the model be maintained at all times so that the agent acting on that model is able to make sound decisions. However, it is not uncommon in the realm of belief-based reasoning for newly acquired information to be inconsistent with the current set of beliefs. For instance, suppose that an agent believes that {\sl Tweety is a bird,} and that {\sl all birds can fly.} This agent subsequently discovers that, {\sl Tweety is a penguin,} and knowing that {\sl penguins can't fly,} will now have to deal with the fact that this new information is inconsistent with what it already believes.

It is clear that when this sort of situation arises, the agent will need to select a set of beliefs to be removed from its knowledge base in order to maintain consistency in its knowledge base when the new belief is inserted, thus allowing it to continue to reason on the basis of its revised knowledge base. However, the question arises: which beliefs should be removed? Even in the above example, the removal of any subset of the beliefs listed will make the new information that {\sl Tweety is a penguin} consistent, so the question of which belief or set of beliefs to remove has no clear answer.

Alchourr\'on, G\"ardenfors and Makinson \cite{agm,AMa,AMb} investigated a set of rationality postulates regarding theory change operators. These postulates appear to capture much of what is required of an ideal rational system of theory change. In particular, they developed three primary classes of theory change operators: expansion, contraction and revision. Expansions are deterministic, and can be expressed set-theoretically, but it is well known that the logical properties of a body of information are not strong enough to uniquely determine a revision or contraction operator. Consequently, some sort of preference order needs to be imposed on the beliefs in the agent's knowledge base. This approach to Belief Revision has come to be known as the AGM paradigm.

Two common models used to impose the preference order required by the AGM paradigm on systems of beliefs are {\sl epistemic entrenchment orderings} \cite{gm} and {\sl systems of spheres} \cite{Grove}. An epistemic entrenchment ordering ranks the sentences in the knowledge base, whereas a system of spheres is a total preorder of the set of possible world states. SATEN uses the epistemic entrenchment representation and uses the services of a full first order theorem prover implemented in Java \cite{VADER}. Revision of a system of spheres can be implemented  
via a standard Database Management System \cite {IIS}.

The implementation of revision in SATEN is steeped in the AGM paradigm \cite{agm}.
G\"ardenfors and Makinson's \cite{gm} showed that an epistemic entrenchment can uniquely determine how an agent will react to the pressures of impinging information. Their construction is an elegant theoretical result, and to use it as the foundation for a computer-based implementation of belief revision the following problems need to be addressed. 

\begin{itemize}
\item
Epistemic entrenchments order all the beliefs in the underlying language. The underlying language may be infinite in size, therefore a finite representation for an epistemic entrenchment is required. 

\item A new knowledge base, rather than a new epistemic entrenchment ordering is constructed. However from a practical point of view developing a new entrenchment ranking is imperative because an implementation must have the capacity to perform iterated revision. 

\end{itemize}

An implementation of belief revision must propagate  an epistemic entrenchment ordering (or more precisely its finite representation).  
SATEN overcomes the first problem using a finite partial entrenchment ranking \cite {JELIA,IJCAI97} as the underlying representation scheme, and it overcomes the second problem using transmutations \cite{Spohn,KR94,IJCAI97}.

\section{Applying the System}
The user specifies a prioritised knowledge base. We have found that for most applications the user-specified ranking can be otained in a natural way during the analysis and design phase \cite{applications}. 

The following  guidelines for designing a ranking were developed in \cite{IJCAI97}:

\begin{itemize}
\item Important information should be made explicit in the ranking
\item Information should be represented in its simplest logical form
\item Maximize the number of ranks, and minimise the number of beliefs at each rank for better performance
\item Rankings should be as irredundant as possible
\item Subsumption at the same rank should be avoided
\item User specifies reasons 
\item Check the contraposition of reasons does not lead to undesired behaviour
\end{itemize}

SATEN can revise first order theory bases.
It uses the services a Java first order theorem prover VADER \cite{VADER} to determine consequence relations, and hence takes worst-case exponential time in the number of formulae in the knowledge base to compute even  propositional revision. However, given Horn input clauses, the system will use a Linear Descent algorithm for its deductions, resulting in linear time efficiency. A number of strategies have been implemented in this system : {\sl standard adjustment}, {\sl maxi-adjustment}, {\sl hybrid adjustment}, {\sl global adjustment}, {\sl linear adjustment}, and {\sl quick adjustment}.

Theory extraction is related to belief revision and uses an extraction operator is a map between epistemic entrenchment orderings which, given an inconsistent ranking, extracts from it some (in some sense) maximal consistent ranking/theory base. Whilst every extraction operator gives rise rather naturally to a corresponding revision operator, it remains unknown whether the converse is true. When SATEN performs a revision, the internal behaviour is  to add the new belief to the base at an appropriate rank, and then to perform a Theory Extraction, before normalising the beliefs to achieve the revised theory base. The process of normalisation converts any ranking into a partial entrenchment ranking.

SATEN is capable of performing both Belief Revision and Theory Extraction on theory bases, and of determining the degree of an arbitrary wff in it's theory base, and can {\sl normalize\/} a ranked theory base. It is also capable of automatically ranking the beliefs in the theory base --- the user has the option to view the total order as a ranking (with the (0,1)-values of the entrenchments suppressed) although the rankings are always represented internally with (0,1)-values. Finally, the user may specify whether new beliefs added to the base without a rank should, by default, be placed at the top of the ranking, or at the bottom. This feature can be used when the rank of new information is unknown. The most conservative approach would place the new information without rank at the both of the ranking.

\subsection{Input Format} SATEN accepts any Predicate Calculus wff's as
input.  Variables in the predicate calculus must begin with a capital
letter, and contain only alphanumeric symbols and underscores (\_).
Constants have the same format as variables, but must begin with a lower
case letter. Predicate and function names must comply with the same format
as variables and constants respectively, but must be followed by a
parenthesized list of arguments separated by commas. 

Input should be in fully parenthesised logic syntax and should not contain
spaces:

\begin{itemize} 
\item Negation is a minus sign (-).
\item Implication is an arrow composed of a minus sign and a greater than sign ($-\hskip -4pt >$). Note no space between the two symbols.
\item Conjunction is an ampersand (\&).
\item Disjunction is a pipe ($\vert$).
\item The universal quantifier is an asterisk (*).
\item The existential quantifier is an exclamation mark (!).
\item Constants must begin with lower-case letters.
\item Variables with upper-case letters.
\item Functions start with lower-case.
\item Propositional statements start with lower-case.
\item First Order Predicates begin with upper-case.
\end{itemize}

Expression grouping is with parentheses, as is function (and predicate) argument list demarkation. User input should not include any double underscores (\_\_) as these are used by the program to tag system-generated variables and skolem functions.
\vskip 6 pt

\noindent Example input format:
\vskip 3 pt
 *X(Psychopathic(X)$\vert$ Emotional(X))     

-null     

hopes\&dreams     

*X(!Y(Mother(X,Y))) 

 !Z(EatsChocolate(Z)$-\hskip -4pt >$Happy(Z))

 *Y(-Income(Y)$-\hskip -4pt >$-Loan(Y))

\hskip 12pt
We can encode "
``{\sl For all $X$ there exists a $Y$ such that $Y$ is the mother of $X$, and for all $Y$, if $Y$ is the daughter of Ben's maternal grandmother, then $Y$ is Ben's mother, or $Y$ is Ben's aunt.\/}'' using the following sentence: 

\[
(*X(!Y(M(X,Y)))\&(*Y(D(MGM(ben),Y) -\hskip -4pt > \]
\vskip -20pt
\[ \hskip 2cm (M(ben,Y)|A(ben,Y)))
\]
where $X$ and $Y$ are variables, $M(X,Y)$ is a predicate of two variables with name $M$, $MGM(X)$ is a function of one variable with name $MGM$, and $ben$ is a constant. 

\subsection{Strategies}
The {\sl standard adjustment} strategy finds the largest cut of the inputs to SATEN that is consistent with the incomming information, and keeps this as the result of the revision. This is quite fast, but removes alot of beliefs from the theory base in comparison to the other strategies because as shown in \cite{IJCAI97} the independence of beliefs must be explicitly specified in the ranking.

The {\sl maxi-adjustment} strategy looks at each individual rank in the belief set, and removes the union of all subsets $S$ of the beliefs on that rank having the properties
\begin{enumerate}
\item $S$ is inconsistent with the union of the incomming information and all the information ranked higher in the base than $S$; and
\item no proper subset of $S$ satisfies $1$.
\end{enumerate}

This strategy keeps many of the beliefs in the theory base, but is also quite slow as it has to examine all subsets of each rank. Performance is improved if the agent is more discerning. our algorithm is anytime \cite{IJCAI97} so the longer the agent has to revise the better the result.

The {\sl hybrid adjustment} strategy represents a combination of the standard adjustment and maxi-adjustment strategies. When revising with $a$, hybrid adjustment first finds every belief $b$ in the base such that $-a|b$ is in the base, and is at the same rank as $-a$ (this is the adjustment step --- it is removing every belief that ``explicitly" fails to be a consequence of the cut above $-a$), and then performs a maxi-adjustment on what is left. So in other words maxi-adjustment is used to recoup as much of the ranking as the time set aside for the revision permits.

This strategy is comparable to maxi-adjustment with regard to the number of beliefs retained, and though somewhat faster on average, has the same worst-case time expenditure as the maxi-adjustment procedure.

The {\sl global adjustment} strategy behaves exactly as maxi-adjustment does, but acting under the assumption that all beliefs have the same rank. This strategy is quite slow, and is not as parsimonious in what it removes as maxi-adjustment, but is highly intuitive with respect to the set of beliefs removed (without reference to their ranks). It comes as close as possible to eliminating the need for a ranking.

The {\sl linear adjustment} strategy \cite{Nebel}looks at the successive cuts of the theory-base and, at every rank at which the cut is inconsistent, removes the entire rank from the base. This is quite time-efficient, and keeps more beliefs in the base than standard adjustment, but, unlike hybrid adjustment and maxi-adjustment, is fairly ad-hoc about the beliefs to be kept in the theory base. 

The {\sl quick adjustment} strategy is an attempt to capture the judiciousness of maxi-adjustment without the associated time expenditure. Once a given rank of beliefs is discovered to cause a contradiction, quick adjustment begins adding beliefs from the left end of that rank until a contradiction is discovered. It then removes these beliefs, one at a time, keeping track of which ones eliminate the contradiction when they are removed. It is these ``culprit'' beliefs that are then removed from the theory base. This process is then iterated.
This strategy has the advantage of improving performance by reducing the
amount of processing
at each
rank, and also emulates the output of maxi-adjustment much of the time. However, its behaviour depends upon the order of the inputs on a given level of the theory base, and hence is somewhat ad hoc. The system could be easily extended to generate more than one ranking and used to perform credulous reasoning. Instead of randomly choosing an alternative, all alternative revised rankings could be created.

\subsection{Other Options} SATEN implements three additional (optional)
strategies that interact with the various Belief Revision strategies
outlined above. These strategies are {\sl subsumption removal,
recovery,\/} and {\sl nuclear revision.\/} Of these three, the second two
are appropriate with any belief revision strategy, but subsumption removal
cannot be used in conjunction with the standard adjustment strategy.
This is the only type of nuclear revision currently implemented. Other
approaches like attributing half-lifes to individual beliefs will be
available in the next version of SATEN.

Using {\sl subsumption removal\/} ensures that any strategy which must choose from a set of beliefs to remove will preferrentially remove beliefs that are subsumed by the incomming belief. For example, maxi-adjustment will look amongst the set of beliefs it was going to remove for the subset that are subsumed by the incomming belief. If there are none, it will proceed as before, but if there are some, it will throw them away first, and then check again for consistency. If the result is still not consistent, the maxi-adjustment algorithm is run again on the remaining beliefs.

The {\sl recovery\/} option ensures that the recovery postulate is satisfied by the revisions. This is achieved by, rather than removing a belief $b$, replacing it with the disjunction $b|a$ where $a$ is the new belief. The only time we do not do this is if $b|a$ is a tautology.

The concept of {\sl nuclear revision} models the fact that beliefs that are not constantly reinforced tend to become less substantial to an intelligent agent. When nuclear revision is enabled, the user will be asked for a half-life. This is the constant factor by which the entrenchment of every belief in the base drops after each operation on the theory base. Thus, if the half-life were set at $0\cdot5$, then the entrenchment of each belief in the theory base would be halved after each operation.

\subsection{Examples}
SATEN has a substantial example menu intended to indicate the correct input form for SATEN, and demonstrate SATEN's capabilities, but also to illustrate the differences between the various belief revision strategies that SATEN implements. These examples are broken into three categories: propositional, predicate, and contrast of strategies.

The first of these indicates the correct input and behaviour of SATEN when presented with propositional inputs only. Under these conditions, SATEN tends to be somewhat more efficient than in the predicate case for the reasons outlined in the VADER hyperbook in the section on performance issues.

The second set of examples indicate SATEN's behaviour with the more general input form of the Predicate Calculus, specifically, its behaviour when quantifiers and variables are involved. The differences are essentially at the theorem prover level, and are opaque to SATEN itself.

The third set of examples are designed to indicate differences in behaviour between the different revision strategies. This is done for two reasons: the first is to highlight the salient points that make each strategy desirable or undesirable in comparison to the others; the second, which is, to some degree subsumed by the first, is to demonstrate that there actually {\it is\/} a difference between each pair of strategies, and that these differences do not require overly sophisticated examples to demonstrate --- in particular, propositional calculus is sufficient, and it is possible to get pairwise distinct behaviour from the strategies on an example with just nine beliefs and four distinct ranks.

\bibliography{nmr2000}
\bibliographystyle{bibaaai}

\end{document}